\documentclass[a4paper,10pt]{article}
\usepackage[headsep=3mm,a4paper,inner=11mm,outer=11mm,top=6mm,includehead,bottom=16mm,heightrounded]{geometry}

\usepackage{mathptmx}       
\usepackage{helvet}         
\usepackage{courier}        
\usepackage{type1cm}        
\usepackage{makeidx}         
\usepackage{graphicx}        
\usepackage[all]{xy}     
\usepackage{multicol}        
\usepackage[bottom]{footmisc}
\usepackage{multirow}
\usepackage{tikz}

\usepackage{color}
\usepackage{epstopdf}
\usepackage{fouridx} 
\usepackage{amsmath}
\usepackage{url}
\usepackage{subfigure}

\DeclareMathOperator*{\argmax}{arg\,max}
\DeclareMathOperator*{\argmin}{arg\,min}

\usetikzlibrary{positioning}
\tikzset{%
	every neuron/.style={
		circle,
		draw,
		minimum size=1cm
	},
	neuron missing/.style={
		draw=none, 
		scale=4,
		text height=0cm,
		execute at begin node=\color{black}$\dots$
	},
}

\makeatletter
\def\@citex[#1]#2{\leavevmode
  \let\@citea\@empty
  \@cite{\@for\@citeb:=#2\do
    {\@citea\def\@citea{,\penalty\@m\ }%
\edef\magic##1{\let##1\expandafter\noexpand\csname bibalias@\@citeb\endcsname}%
\magic\tmp \ifx\tmp\relax\else \let\@citeb\tmp\fi
     \edef\@citeb{\expandafter\@firstofone\@citeb\@empty}%
     \if@filesw\immediate\write\@auxout{\string\citation{\@citeb}}\fi
     \@ifundefined{b@\@citeb}{\hbox{\reset@font\bfseries ?}%
       \G@refundefinedtrue
       \@latex@warning
         {Citation `\@citeb' on page \thepage \space undefined}}%
       {\@cite@ofmt{\csname b@\@citeb\endcsname}}}}{#1}}
\def\bibalias#1#2{\expandafter\def\csname bibalias@#1\endcsname{#2}}
\makeatother

\bibalias{dredze2010multiDomain}{DredzeML10Multidomain}
\bibalias{Evgeniou2004}{EvgeniouSIGKDD04Regularized}
\bibalias{Argyriou2008}{ArgyriouML08Convex}
\bibalias{daume2012gomtl}{KumarICML12Learning}
\bibalias{MuandetBS13}{MuandetICML13Domain}
\bibalias{Ji2009tracenormmtl}{JiICML09Accelerated}
\bibalias{larochelle2008zerodata}{LarochelleAAAI08Zerodata}
\bibalias{socher2013zslCrossModal}{SocherNIPS13Zeroshot}
\bibalias{Palatucci_2009_6459}{PalatucciNIPS09Zero}
\bibalias{DeViSE}{FromeNIPS13Devise}
\bibalias{Blitzer_zero-shotdomain}{BlitzerTR09Zeroshot}
\bibalias{icml2013_romera-paredes13}{RomeraParedesICML13Multilinear}
\bibalias{MemisevicH07}{MemisevicCVPR07Unsupervised}
\bibalias{tensorflow2015}{Tensorflow2015}
\bibalias{caltech256}{GriffinTR07Caltech}
\bibalias{Bay2008surf}{BayECCV06Surf}
\bibalias{hoffman2014continuousDA}{HoffmanCVPR14Continuous}
\bibalias{ZhengICIP2011Robust}{ZhengICIP11Robust}
\bibalias{YangCVPR16Domains}{YangCVPR16Multivariate}
\bibalias{YangX16tensor}{YangX16Deep}
\bibalias{Hitchcock1927Expression}{HitchcockJMP27Expression}
\bibalias{Tuck1966Some}{TuckPsy66Some}
\bibalias{Oseledets2011Tensor}{OseledetsJOS11TensorTrain}
\bibalias{sigaud2015gatedInventory}{SigaudX15Gated}
\bibalias{AlainO13}{DroniouICML13Gated}

\begin{document}
	
\title{Unifying Multi-Domain  Multi-Task Learning:\\ Tensor and Neural Network Perspectives}
\author{
Yongxin Yang, Timothy M. Hospedales\\
\small Queen Mary, University of London\\
\small \url{yongxin.yang@qmul.ac.uk}, \url{t.hospedales@qmul.ac.uk}
}
\date{}
\maketitle

\abstract{Multi-domain learning aims to benefit from simultaneously learning across several different but related domains. In this chapter, we propose a single framework that unifies multi-domain learning (MDL) and the related but better studied area of multi-task learning (MTL). By exploiting the concept of a \emph{semantic descriptor} we show how our framework encompasses various classic and recent MDL/MTL algorithms as special cases with different semantic descriptor encodings. As a second contribution, we present a higher order generalisation of this framework, capable of simultaneous multi-task-multi-domain learning. This generalisation has two mathematically equivalent views in multi-linear algebra and gated neural networks respectively. Moreover, by exploiting the semantic descriptor, it provides neural networks the capability of zero-shot learning (ZSL), where a classifier is generated for an unseen class without any training data; as well as zero-shot domain adaptation (ZSDA), where a model is generated for an unseen domain without any training data. In practice, this framework provides a powerful yet easy to implement method that can be flexibly applied to MTL, MDL, ZSL and ZSDA.}

\section{Introduction}

The multi-domain setting arises when there is data about a task in several different but related domains. For example in visual recognition of an object when viewed with different camera types. Multi-domain learning (MDL) models \cite{dredze2010multiDomain,DaumeACL07Frustratingly,YangICLR15Unified} aim to learn a cross-domain parameter sharing strategy that reflects the domains' similarities and differences. Such selective parameter sharing aims to be robust to the differences in statistics across domains, while exploiting data from multiple domains to improve performance compared to learning each domain separately.

In this chapter we derive a general framework that encompasses MDL and MTL from both neural network and tensor-factorisation perspectives.  Many classic and recent MDL/MTL algorithms can be understood by  the assumptions about the cross domain/task sharing structure encoded in their designs. E.g., the assumption that each task/domain's model is a linear combination of a global and a task-specific parameter vector \cite{Evgeniou2004, DaumeACL07Frustratingly}. Our framework includes these as special cases corresponding to specific settings of a \emph{semantic descriptor vector} parametrising tasks/domains \cite{YangICLR15Unified}. This vector can be used to recover existing models from our framework, but more generally it allows one to relax the often implicit assumption that domains are atomic/categorical entities, and exploit available \emph{metadata} about tasks/domains to guide sharing structure for better MDL/MTL \cite{YangICLR15Unified,YangCVPR16Domains}. For example, in surveillance video analysis, exploiting the knowledge of the time of day and day of week corresponding to each domain for better MDL. Finally, the idea of a semantic task/domain descriptor, allows our framework to go beyond the conventional MDL/MTL setting, and address both zero-shot learning \cite{YangICLR15Unified} and zero-shot domain adaptation \cite{YangICLR15Unified,YangCVPR16Domains} -- where a model can be deployed for a new task/domain without any training data, solely by specifying the task/domain's semantic descriptor metadata.

\paragraph{\textbf{Multi-Domain versus Multi-Task Learning} }

The difference between domains and tasks can be subtle, and some multi-domain learning problems can be addressed by  methods  proposed for multi-task learning and vice-versa. However, to better understand this work, it is useful to distinguish them clearly. Domains refer to multiple datasets addressing the same task, but with differing statistical bias. For example camera type for object recognition; time of day or year for surveillance video analysis; or more subtle human biases in data collection \cite{TorralbaCVPR11Unbiased}. Tasks, on the other hand would refer to different object categories to recognise. In other words, a task change affects the output label-space of a supervised learning problem, while a domain change does not. 

A classic benchmark with multiple domains is the Office dataset \cite{SaenkoECCV10Adapting}. It contains images of the same set of   categories (e.g., mug, laptop, keyboard) from three data sources (Amazon website, webcam, and DSLR). In this context, multi-task learning could improve performance by sharing information about how to recognise keyboard and laptop; while multi-domain learning could improve performance by sharing knowledge about how to recognise those classes in Amazon product versus webcam images. Some problems can be interpreted as either setting. E.g., in the School dataset \cite{Argyriou2008} the goal is to predict students' exam scores based on their characteristics. This dataset is widely used to evaluate MTL algorithms, where students from different schools are grouped into different tasks. However, one can argue that school groupings are better interpreted as domains than tasks. 

As a rule of thumb, multi-domain learning problems occur when a model from domain A could be directly applied to domain B albeit with reduced performance; while multi-task learning problems occur where a model for task A can not meaningfully be applied to task B because their label-spaces are fundamentally different. In some problems, the multi-domain and multi-task setting occur simultaneously. E.g., in the Office dataset there are both multiple camera types, and multiple objects to recognise. Few existing methods can deal with this setting. MTL methods break a multi-class problem into multiple 1-v-all tasks and share information across tasks \cite{Argyriou2008,daume2012gomtl}, while MDL methods typically deal with a single-output problem in multiple domains \cite{DaumeACL07Frustratingly}. Our higher order generalisation addresses simultaneous multi-domain multi-task learning as required by problems such as the one posed by Office.

\paragraph{\textbf{Relation to Domain Adaptation and Domain Generalisation}}

Dataset bias/domain-shift means that models trained on one domain often have weak performance when deployed in another domain. The community has proposed two different approaches to alleviate this: (i) Domain adaptation (DA): calibrating a pre-trained model to a target domain using a limited amount of labelled data -- supervised DA \cite{SaenkoECCV10Adapting}, or unlabelled data only -- unsupervised DA \cite{GongCVPR12Geodesic}, or both --semi-supervised DA \cite{LiPAMI14Learning}. (ii) Domain generalisation (DG): to train a model that can is insensitive to domain bias, e.g., learning domain invariant features \cite{MuandetBS13}.

The objective of multi-domain learning is different from the mentioned domain adaptation and domain generalisation. MDL can be seen as a bi-directional generalisation of DA with each domain benefiting the others, so that all domains have maximal performance; rather than solely transferring knowledge from source $\to$ target domain as in DA. In its conventional form MDL does not overlap with DG, because it aims to improve the performance on the set of given domains, rather than address a held out domain. However, our zero-shot domain adaptation extension of MDL, relates to DG insofar as aiming to address a held-out domain. The difference is that we require a semantic descriptor for the held out domain, while DG does not. However where such a descriptor exists, ZSDA is expected to outperform DG.

\section{Methodology -- Single Output Models} \label{sec:singleouput}

We start from the linear model assumption made by most popular MTL/MDL methods \cite{Argyriou2008,Ji2009tracenormmtl} -- i.e., that a domain/task corresponds to a univariate linear regression or binary classification problem. The case where a domain/task needs more than that (e.g., multi-class problem) will be discussed in Section~\ref{sec:multileouput}.  We also assume the domains/tasks are homogeneous, i.e., different domains/tasks have model representations and instance feature vectors of the same size.

\subsection{Formulation: Vector-Matrix Inner Product}

Assume we have M domains (tasks), and the $i$th domain has $N_i$ instances. We denote the $D$-dimensional feature vector of $j$th instance in $i$th domain (task) and its associated $B$-dimensional semantic descriptor\footnote{Note that, in any multi-domain or multi-task learning problem, all  instances are at least implicitly associated with a semantic descriptor indicating their domain (task). }\footnote{All vectors (feature  $x$, weights  $w$, and domain descriptor $z$) are by default column vectors.} by the pair
$\{\{x^{(i)}_j,z^{(i)}\}_{j=1}^{N_i}\}_{i=1}^M$
 and the corresponding label 
 $\{\{y^{(i)}_j\}_{j=1}^{N_i}\}_{i=1}^M$.
All the weight vectors $w^{(i)}$ for each domain (task) $i$ can be stacked into a single weight matrix $\tilde{W}$. Without loss of generality, we to minimise the empirical risk for all domains (tasks),
\begin{equation}
\argmin_{\tilde{W}}\frac{1}{M}\sum_{i=1}^{M}\bigg(\frac{1}{N_i}\sum_{j=1}^{N_i}\ell\Big(\hat{y}^{(i)}_j,y^{(i)}_j\Big)\bigg).\label{eq:obj1}
\end{equation}
\noindent Here $\ell(\hat{y}, y)$ is a loss function depending on the predicted $\hat{y}$ and true label $y$. For the $i$th domain/task, the linear model $w^{(i)}$ makes predictions as:

\begin{equation}
\hat{y}^{(i)}_j = x^{(i)}_j \cdot w^{(i)} = {x^{(i)}_j}^T w^{(i)}.
\label{bc_eq1}
\end{equation}

Now we introduce a key idea in our work: rather than being learned directly, each task/domain's model $w^{(i)}$ is \emph{generated} by a linear function $f(\cdot)$ of its descriptor, {\color{black} termed the weight generating function,} 

\begin{equation}
w^{(i)} = f(z^{(i)}) = Wz^{(i)}.
\label{bc_eq2}
\end{equation}

\noindent That is, the linear model for the $i$th domain/task is produced by its $B$-dimensional semantic descriptor hitting a $D\times B$ matrix $W$. If all semantic descriptors are stacked (as columns) into matrix $Z$, we have

\begin{equation}
\tilde{W} = f(Z) = WZ 
\label{bc_eq2all}
\end{equation}

Rather than learning the per-domain models $\tilde{W}$ independently as in Eq.~\ref{eq:obj1}, we now learn the {\color{black} \emph{weight generating function}} parametrised by matrix $W$ instead. By substituting Eq.~\ref{bc_eq2} into Eq.~\ref{bc_eq1}, we can re-write the prediction for the $i$th task/domain into a bilinear form,
\begin{equation}
\hat{y}^{(i)}_j = {x^{(i)}_j}^T W z^{(i)}.
\label{bc_eq3}
\end{equation}

This general formulation encompasses many MTL/MDL methods. As we will see, the two key axes of design space on which existing  MTL/MDL algorithms differ are the encoding of descriptors $z$ and decomposition or regularisation of matrix $W$. We will discuss these in the following two sections.

\subsection{Semantic Descriptor Design}

\noindent\textbf{One-hot encoding $z$}\quad
In the simplest scenario  $z$ is a one-hot encoding vector that \emph{indexes} domains (Fig.~\ref{fig:dd1hot}). The model generation function $f(z^{(i)})$ then just \emph{selects} one column from the matrix $W$. For example, $z^{(1)} = [1, 0, 0]^T$, $z^{(2)} = [0, 1, 0]^T$, $z^{(3)} = [0, 0, 1]^T$ if there are $M=3$ domains/tasks. In this case, the length of the descriptor and the number of unique domains (tasks) are equal $B = M$, and the stack of all $z^{i}$ vectors (denoted $Z=[z^{(1)}, z^{(2)}, \dots]$) is an $M\times M$ identity matrix. Learning the `weight generating function' is then equivalent to independently learning per-domain weights.

\begin{figure}[t]
\minipage{0.48\textwidth}
\begin{equation*}
\resizebox{0.95\hsize}{!}{%
$Z = 
\begin{bmatrix}
& \text{Domain-1} & \text{Domain-2} & \text{Domain-3} \\
\text{Index-1} &1 & 0 & 0 \\
\text{Index-2} &0 & 1 & 0 \\
\text{Index-3} &0 & 0 & 1 \\
\end{bmatrix}$%
}
\end{equation*}
\endminipage
~~~
\minipage{0.48\textwidth}
\begin{equation*}
\resizebox{0.95\hsize}{!}{%
$Z = 
\begin{bmatrix}
& \text{Domain-1} & \text{Domain-2} & \text{Domain-3} \\
\text{Index-1} & 1 & 0 & 0 \\
\text{Index-2} & 0 & 1 & 0 \\
\text{Index-3} & 0 & 0 & 1  \\
\text{Shared} & 1 & 1 & 1 
\end{bmatrix}$%
}
\end{equation*}
\endminipage
\caption{Domain descriptor for categorical/atomic domains. One-hot encoding (left), and one-hot with constant encoding (right).}\label{fig:dd1hot}
\end{figure}

\vspace{0.2cm}\noindent\textbf{One-hot encoding $z$ with a constant}\quad
A drawback of one-hot encoding is that the $z^{i}$ are orthogonal to each other, which suggests that all domains/tasks are independent -- there is no cross domain/task information sharing. To encode an expected sharing structure of an underlying commonality across all  domains/tasks, an alternative approach to constructing $z$ is to append a constant term after the one-hot encoding. For the case of $M=3$, we might have $z^{(1)} = [1, 0, 0, 1]^T$, $z^{(2)} = [0, 1, 0, 1]^T$, $z^{(3)} = [0, 0, 1, 1]^T$. Fig.~\ref{fig:dd1hot} shows the resulting $B\times M$ matrix $Z$ (in this case, $B=M+1$). The prediction of task (domain) $i$ is given as $\hat{y}^{(i)}={x^{(i)}}^TWz^{(i)}={x^{(i)}}^T(w^{(i)}+w^{(4)})$, i.e., the sum of a task/domain specific and a globally shared prediction. Learning the weight generator corresponds to training both the local and shared predictors. This approach is implicitly used by some classic MDL/MTL  algorithms \cite{Evgeniou2004, DaumeACL07Frustratingly}.

\begin{figure}[t]
\minipage{0.48\textwidth}
\begin{equation*}
\resizebox{0.95\hsize}{!}{%
$Z = 
\begin{bmatrix}
& \text{Domain-1} & \text{Domain-2} & \text{Domain-3} & \text{Domain-4} \\
\text{A-1-B-1} & 1 & 0 & 0 & 0 \\
\text{A-1-B-2} & 0 & 1 & 0 & 0 \\
\text{A-2-B-1} & 0 & 0 & 1 & 0 \\
\text{A-2-B-2} & 0 & 0 & 0 & 1
\end{bmatrix}$%
}
\end{equation*}
\endminipage
~~
\minipage{0.48\textwidth}
\begin{equation*}
\resizebox{0.95\hsize}{!}{%
$Z = 
\begin{bmatrix}
& \text{Domain-1} & \text{Domain-2} & \text{Domain-3} & \text{Domain-4} \\
\text{A-1} & 1 & 1 & 0 & 0 \\
\text{A-2} & 0 & 0 & 1 & 1 \\
\text{B-1} & 1 & 0 & 1 & 0 \\
\text{B-2} & 0 & 1 & 0 & 1
\end{bmatrix}$%
}
\end{equation*}
\endminipage
\caption{Example domain descriptor for domains with multiple factors. One-hot encoding (left). Distributed encoding (right).} \label{fig:ddDistrib}
\end{figure}

\vspace{0.2cm}\noindent\textbf{Distributed encoding $z$}\quad
In most studies of MDL/MTL, domain or task is assumed to be an atomic category which can be effectively encoded by the above indexing approaches. However more structured domain/task-metadata is often available, such that a domain (task) is indexed by multiple factors (e.g., time: day/night, and date: weekday/weekend, for batches of video surveillance data). Suppose two categorical variables (A,B) describe a domain, and each of them has two states (1,2), then at most four distinct domains exist. Fig.~\ref{fig:ddDistrib}(left) illustrates the semantic descriptors for 1-hot encoding. However this encoding does not exploit the sharing cues encoded in the metadata (e.g., in the surveillance example that day-weekday should be more similar to day-weekend and night-weekday than to night-weekend). Thus we propose to use a distributed encoding of the task/domain descriptor (Fig~\ref{fig:ddDistrib}(right)). Now  prediction weights are given by a linear combination of $W$'s columns given by the descriptor, and learning the weight generating function means learning weights for each factor (e.g., day, night, weekday, weekend). We will demonstrate that the ability to exploit structured domain/task descriptors where available, improves information sharing compared to existing MTL/MDL methods in later experiments.

\subsection{Unification of Existing Algorithms}

The key intuitions of various MDL/MTL methods are encoded in constraints on $W$ in Eq.~\ref{bc_eq2} and/or encodings of descriptors $z$. Besides enforcing local and shared components \cite{Evgeniou2004, DaumeACL07Frustratingly} (which we interpret as a one-hot+constant descriptor) many studies achieve information sharing via enforcing low-rank structure on $\tilde{W}$ (the $D\times T$ stack of weight vectors for each task). 
Popular choices for modelling $\tilde{W}$ are: (i) add regularisation that encourages $\tilde{W}$ to be a low-rank matrix (ii) explicit low rank factorisation $\tilde{W}=\tilde{P}\tilde{Q}$, and optionally placing some constraint(s) on $\tilde{P}$ and/or $\tilde{Q}$.
We assume our weight generating function, the $D\times B$ matrix $W$, is replaced by the dot-product of two factor matrices $P$ and $Q$ ($D\times K$ and $K\times B$ respectively). Thus Eq.~\ref{bc_eq2} becomes
\begin{equation}
w^{(i)} = f(z^{(i)}) = PQz^{(i)}
\label{bc_eq2pq}
\end{equation}

\noindent That is, we generate specific weights for prediction by combining the task/domain descriptor $z^{(i)}$ with matrices $P$ and $Q$; and learning corresponds to fitting $P$ and $Q$.

\begin{table}[t]
\begin{center}
\scalebox{0.9}{
\begin{tabular}{c c c c c c}
\hline  & $Z$ & $P$ & Norm on $P$ & $Q$ & Norm on $Q$ \\ \hline
RMTL \cite{Evgeniou2004} \& FEDA \cite{DaumeACL07Frustratingly} & $\begin{bmatrix}1 & 0 & 0 \\0 & 1 & 0 \\0 & 0 & 1 \\ 1 & 1 & 1  \end{bmatrix}$ & Identity & None & $\begin{bmatrix}| & | & | & | \\v_1 & v_2 & v_3 & w_0\\| & | & | & |  \end{bmatrix}$ & None \\  
MTFL \cite{Argyriou2008} & $\begin{bmatrix}1 & 0 & 0 \\0 & 1 & 0 \\0 & 0 & 1  \end{bmatrix}$ & Identity & None & W & $\ell_{2, 1}$ Norm \\ 
TNMTL \cite{Ji2009tracenormmtl} & $\begin{bmatrix}1 & 0 & 0 \\0 & 1 & 0 \\0 & 0 & 1  \end{bmatrix}$ & Identity & None & W & Trace Norm \\ 
GO-MTL \cite{daume2012gomtl} & $\begin{bmatrix}1 & 0 & 0 \\0 & 1 & 0 \\0 & 0 & 1 \end{bmatrix}$ & L & Frobenius & S & Entry-wise $\ell_1$ \\ 
\hline
\end{tabular}}
\caption{Existing methods as special cases of our framework. Each column of the $Z$ matrices corresponds to the assumed domain/task  descriptor ($z^{(i)}$) encoding. The notion used in the table is the same as the original paper, e.g., $P$  here is analogous to $L$ in \cite{daume2012gomtl}.}\label{tab:specialCase}
\end{center}
\end{table}

A variety of existing algorithms\footnote{RMTL: \textbf{R}egularized \textbf{M}ulti--\textbf{T}ask \textbf{L}earning, FEDA: \textbf{F}rustratingly \textbf{E}asy \textbf{D}omain \textbf{A}daptation, MTFL: \textbf{M}ulti--\textbf{T}ask \textbf{F}eature \textbf{L}earning, TNMTL: \textbf{T}race-\textbf{N}orm \textbf{M}ulti--\textbf{T}ask \textbf{L}earning, and GO-MTL: \textbf{G}rouping and \textbf{O}verlap for \textbf{M}ulti--\textbf{T}ask \textbf{L}earning} are then special cases of our general framework. We illustrate this via MDL/MTL  with $M=3$ domains/tasks. Observe that RMTL \cite{Evgeniou2004}, FEDA \cite{DaumeACL07Frustratingly}, MTFL \cite{Argyriou2008}, TNMTL \cite{Ji2009tracenormmtl}, and GO-MTL \cite{daume2012gomtl} correspond to asserting specific settings of $Z$, $P$ and $Q$ as shown in Tab~\ref{tab:specialCase}.

\subsection{A Two-sided Network View}
\label{sec:twosidenn}
We now provide an alternative neural network view of the class of methods introduced above. By substituting Eq.~\ref{bc_eq2pq} into Eq.~\ref{bc_eq1}, we see the prediction is given by
\begin{equation}
\hat{y}^{(i)}_j = {x^{(i)}_j}^T P Q z^{(i)}
\label{bc_eq3pq}
\end{equation}

A mathematically equivalent neural network model is illustrated in Fig.~\ref{fig:model}. The introduction of this NN leads to a better understanding of our framework and the set of methods it encompasses. The left-hand side can be understood as a global representation learning network -- $x^T P$, and the right-hand side can be seen as a network that generates a model (the weight vector $z^T Q^T$) for the  task/domain encoded by $z$. This interpretation allows existing MTL/MDL models to be implemented as neural networks --  and conveniently optimised by standard neural network libraries -- by setting inputs $z$ appropriately, and activating appropriate weight regularisation (Table~\ref{tab:specialCase}). However, going beyond existing methods, we exploit the semantic descriptor as an input, which allows information sharing to be guided by domain metadata \cite{YangICLR15Unified} where available. Moreover, as a NN, both sides can go arbitrarily deeper \cite{YangX16tensor}. E.g., the representation learning network can be a full-sized CNN that extracts features from raw images and everything can be trained end-to-end with back propagation.

\begin{center}
\begin{figure}[t]
\centering
\resizebox{0.9\linewidth}{!}{%
\begin{tikzpicture}[x=1.5cm, y=1.5cm, >=stealth]

\foreach \m/\l [count=\y] in {1,2,3,missing,4}
\node [every neuron/.try, neuron \m/.try] (input-\m) at (\y,0) {};

\foreach \m [count=\y] in {1,2,3,missing,4}
\node [every neuron/.try, neuron \m/.try ] (hidden-\m) at (\y,1.5) {};

\foreach \i in {1,...,4}
\foreach \j in {1,...,4}
\draw [->] (input-\i) -- (hidden-\j);

\node [align=right] at (0.25,0) {\huge $x^T$};
\node [align=right] at (0.25,0.75) {\huge $P$};
\draw (0.5,-0.5) -- (5.5,-0.5) -- (5.5,0.5) -- (0.5,0.5) -- (0.5,-0.5);
\draw (0.5,1) -- (5.5,1) -- (5.5,2) -- (0.5,2) -- (0.5,1);
\draw (3,2) -- (3,2.5) -- (5.875,2.5);

\pgfmathsetmacro{\PP}{5.75}

\foreach \m/\l [count=\y] in {1,2,3,missing,4}
\node [every neuron/.try, neuron \m/.try] (input2-\m) at (\PP+\y,0) {};

\foreach \m [count=\y] in {1,2,3,missing,4}
\node [every neuron/.try, neuron \m/.try ] (hidden2-\m) at (\PP+\y,1.5) {};

\foreach \i in {1,...,4}
\foreach \j in {1,...,4}
\draw [->] (input2-\i) -- (hidden2-\j);

\node [align=left] at (\PP+5.75,0) {\huge $z^T$};
\node [align=right] at (\PP+5.75,0.75) {\huge $Q^T$};

\draw (\PP+0.5,-0.5) -- (\PP+5.5,-0.5) -- (\PP+5.5,0.5) -- (\PP+0.5,0.5) -- (\PP+0.5,-0.5);
\draw (\PP+0.5,1) -- (\PP+5.5,1) -- (\PP+5.5,2) -- (\PP+0.5,2) -- (\PP+0.5,1);
\draw (\PP+3,2) -- (\PP+3,2.5) -- (5.875,2.5);

\draw[thick,->] (5.875,2.5) -- (5.875,2.75);

\node [every neuron] (output1) at (5.875,3.075) {\huge $y$};

\end{tikzpicture}
}
\caption{Two-sided Neural Network for Multi-Task/Multi-Domain Learning} \label{fig:model}
\end{figure}
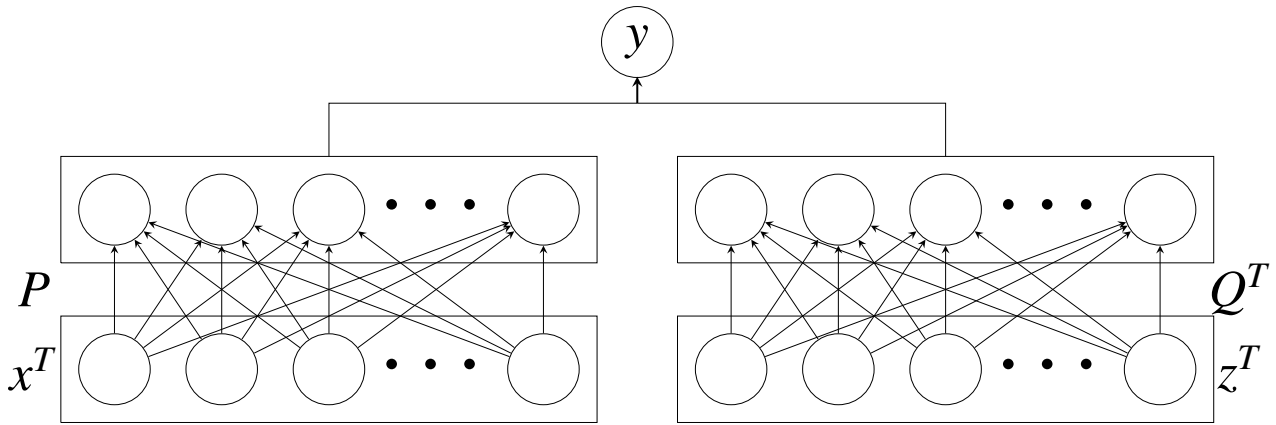
\end{center}

\subsection{Application to Zero-Shot Learning}

\subsubsection{Zero-Shot Recognition} 

Zero-Shot Learning (ZSL) aims to eliminate the need for training data for a particular task. It has been widely studied in areas such as character  \cite{larochelle2008zerodata} and object recognition \cite{LampertCVPR09Learning,socher2013zslCrossModal}. Typically for ZSL, the label space of training and test data are disjoint, so no data has been seen for test-time categories. Instead, test-time classifiers are constructed given some mid-level information. Most existing ZSL methods follow data flows that can be illustrated as either: $X\rightarrow Z \rightarrow Y$ \cite{Palatucci_2009_6459} or $\xymatrix{Z \ar@<0.1pt>@/^0.5pc/[rr] & X \ar[r] & Y}$ \cite{larochelle2008zerodata,DeViSE}  where where $Z$ is some ``semantic descriptor", e.g.,  attributes \cite{LampertCVPR09Learning} or semantic word vectors \cite{socher2013zslCrossModal}.
In our framework ZSL can be achieved via the latter pipeline, implemented by the network in Fig.~\ref{fig:model} \cite{YangICLR15Unified}. By presenting each novel semantic vector $z^{(t)}$ (assuming testing category are indexed by $t$) in turn along with novel category instance $x_*$. Zero-shot recognition for $x_*$ then is given by: $\hat{t}=\argmax_{t} x_*^T P Q z^{(t)}$. In this chapter we focus our experiments on zero-shot domain adaptation (parametrised domains), the interested reader can see \cite{YangICLR15Unified} for experiments applying our framework to zero-shot recognition (parametrised tasks).

\subsubsection{Zero-Shot Domain Adaptation}

Going beyond conventional ZSL, we generalise the notion of zero-shot learning of tasks to zero-shot learning of domains. In this context, zero-shot means no training data has been seen for the target domain prior to testing. The challenge is to construct a good model for a novel test domain based solely on its semantic descriptor \cite{YangICLR15Unified,YangCVPR16Domains}. We denote this problem setting as Zero-Shot Domain Adaptation\footnote{Note that despite the title, \cite{Blitzer_zero-shotdomain} actually considers unsupervised domain adaptation without target domain labels, but \emph{with} target domain data.} (ZSDA)

ZSDA becomes possible with a distributed rather than one-hot encoded domain descriptor, as in practice only a subset of domains is necessary to effectively learn $Q$. Thus a model $w^{(*)}$ suitable for an \emph{unseen} domain can be constructed without any training data -- by applying its semantic descriptor $z^{(*)}$ to the model generation matrix $Q$: $w^{(*)}=Qz^{(*)}$. The generated domain-specific model -- $w^{(*)}$ -- is then then used to make predictions for the re-represented input: $x^T_*Pw^{(*)}$.

\section{Methodology -- Multiple Output Models}\label{sec:multileouput}

Thus far, the final output of each model is a scalar (single output). However for some practical applications, multiple outputs are desirable or required. For example, assume that we have $M=2$ handwriting digit datasets (domains): MNIST and USPS. For any MNIST or USPS image, a $D$-dimensional feature vector is extracted. The task is to classify the image from 0 to 9 and thus we have $D\times C$ ($C=10$) model parameters for each dataset. Therefore, the full model for all digits and datasets should  contain $D\times C \times M$ parameters. 
We denote this setting of multiple domains, each of which has multiple tasks, as multi-domain-multi-task learning. In some recent literature \cite{icml2013_romera-paredes13} a similar setting is named multi-linear multi-task learning.

\subsection{Formulation: Vector-Tensor Inner Product}

The key idea in the previous section was to generate a model vector via a descriptor vector hitting a matrix (Eq.~\ref{bc_eq2}). To adapt this idea for this new setting, we propose

\begin{equation}
W^{(i)} = f(z^{(i)}) = \mathcal{W}\times_3 z^{(i)}
\label{bc_eq2tensor}
\end{equation}
\noindent where $\times_n$ indicates the $n$-mode product of a tensor and vector (this is also referred to as tensor contraction in some studies as it is a generalisation of inner product for vectors and/or matrices). The generated model is now a weight \emph{matrix} $W^{(i)}$ rather than a \emph{vector} $w^{(i)}$. The weight generating function is now parametrised by a third-order tensor $\mathcal{W}$ of size $D\times C \times B$, and it synthesises the model matrix for the $i$th domain by hitting the tensor with its $B$-dimensional semantic descriptor $z^{(i)}$. This is a natural extension: if the required model is a vector (single output), the weight generating function is a matrix (second-order tensor) hits the semantic descriptor $z$ (Eq.~\ref{bc_eq2}); when the required model is a matrix (multiple outputs), the weight generating function is then $z$ hits a third-order tensor.

Given one-hot encoding descriptors $z^{(1)} = [1,0]^T$ and $z^{(2)} = [0,1]^T$ indicating MNIST and USPS respectively. Eq.~\ref{bc_eq2tensor} would just \emph{slice} an appropriate matrix from the tensor $\mathcal{W}$. However alternative and more powerful distributed encodings of $z^{(i)}$ are also applicable. The model prediction from Eq.~\ref{bc_eq3} is then generalised as

\begin{equation}
\hat{y}^{(i)}_j = \mathcal{W} \times_1 {x^{(i)}_j} \times_3 z^{(i)}
\label{bc_eq3tensor}
\end{equation}

\noindent where $\hat{y}^{(i)}_j$ is now a $C$-dimensional vector instead of a scalar as in Eq.~\ref{bc_eq3}. Nevertheless, this method does not provide information sharing in the case of conventional  categorical (1-hot encoded) domains. For this we turn to tensor factorisation next.

\subsection{Tensor (De)composition} 

Recall that the key intuition of many classic (matrix-based) multi-task learning methods is to exploit the information sharing induced by the row-rank factorisation $\tilde{W}=\tilde{P}\tilde{Q}$. I.e., composing the weight matrix for all tasks $\tilde{W}$ from factor matrices $\tilde{P}$ and $\tilde{Q}$. For MDL with multiple outputs, we aim to extend this idea to the factorisation of the weight tensor $\mathcal{W}$. In contrast to the case with matrices, there are multiple approaches to factorising tensors, including CP \cite{Hitchcock1927Expression}, Tucker \cite{Tuck1966Some}, and Tensor-Train \cite{Oseledets2011Tensor} Decompositions.

\subsubsection{CP decomposition}

For a third-order tensor $\mathcal{W}$ of size $D\times C \times B$, the rank-$K$ CP decomposition is:
\begin{eqnarray}
\mathcal{W}_{d,c,b} &=& \sum_{k=1}^{K} U^{(D)}_{k,d} U^{(C)}_{k,c} U^{(B)}_{k,b}\\
%
\mathcal{W} &=& \sum_{k=1}^{K} U^{(D)}_{k,\cdot}\odot U^{(C)}_{k,\cdot} \odot U^{(B)}_{k,\cdot}\label{cp}
\end{eqnarray}
\noindent where $\odot$ is outer product. The factor matrices $U^{(D)}$, $U^{(C)}$, and $U^{(B)}$ are of respective size $K\times D$, $K\times C$, and $K\times B$.

Given a data point $x$ and its corresponding descriptor $z$\footnote{We omit the upper- and lower- scripts for clarity.}, Eq.~\ref{bc_eq3tensor} will produce a $C$-dimensional vector $y$ (e.g., $C=10$ the scores of 10 digits for the MNIST/USPS example). By substituting Eq.~\ref{cp} into Eq.~\ref{bc_eq3tensor} and some reorganising, we obtain

\begin{equation}
y = {U^{(C)}}^T((U^{(D)}x)\circ(U^{(B)}z))
\label{eq:ycp}
\end{equation}

\noindent where $\circ$ is the element-wise product. It also can be written as,

\begin{equation}
y = {U^{(C)}}^T \operatorname{diag}(U^{(B)}z) U^{(D)}x
\end{equation}

\noindent from which we obtain a specific form of the weight generating function in Eq.~\ref{bc_eq2tensor}, which is motived by CP decomposition:

\begin{equation}
W^{(i)} = f(z^{(i)}) = \mathcal{W}\times_{3}z^{(i)} =  {U^{(D)}}^T\operatorname{diag}(U^{(B)}z) U^{(C)}
\label{bc_eq2tensorcp}
\end{equation}

It is worth mentioning that this formulation has been used in the context of gated neural networks \cite{sigaud2015gatedInventory}, such as Gated Autoencoders \cite{AlainO13}. However, \cite{AlainO13} uses the technique to model the relationship between two inputs (images), while we exploit it for knowledge sharing in multi-task/multi-domain learning.

\subsubsection{Tucker decomposition}

Given the same sized tensor $\mathcal{W}$, Tucker decomposition outputs a core tensor $\mathcal{S}$ of size $K_D\times K_C \times K_B$, and $3$ matrices $U^{(D)}$ of size $K_D \times D$, $U^{(C)}$ of size $K_C \times C$, and $U^{(B)}$ of size $K_B \times B$, such that,
\begin{eqnarray}
\mathcal{W}_{d,c,b}  &=&  \sum_{k_D=1}^{K_D}\sum_{k_C=1}^{K_C}\sum_{k_B=1}^{K_B} \mathcal{S}_{k_D,k_C,k_B}{U}^{(D)}_{k_D,d}{U}^{(C)}_{k_C,c}{U}^{(B)}_{k_B,b}\\
\mathcal{W} &=&  \mathcal{S} {\times}_1 {U}^{(D)} {\times}_2 {U}^{(C)} {\times}_3 {U}^{(B)}\label{tucker}
\end{eqnarray}

\noindent Substituting Eq.~\ref{tucker} into Eq.~\ref{bc_eq3tensor}, we get the prediction for instance $x$ in domain/task $z$

\begin{equation}
y = ((U^{(D)}x)\otimes(U^{(B)}z))\mathcal{S}_{(2)}^T U^{(C)}
\label{eq:ytucker}
\end{equation}  

\noindent where $\otimes$ is Kronecker product. $\mathcal{S}_{(2)}$ is the mode-2 unfolding of $\mathcal{S}$ which is a $K_C \times K_D K_B$ matrix, and its transpose $\mathcal{S}_{(2)}^T$ is a matrix of size $K_D K_B \times K_C$.

This formulation was used by studies of Gated Restricted Boltzmann Machines (GRBM) \cite{MemisevicH07} for similar image-transformation purposes as \cite{AlainO13}. The weight generating function (Eq.~\ref{bc_eq2tensor}) for Tucker decomposition is

\begin{equation}
	W^{(i)} = f(z^{(i)}) = \mathcal{W}\times_{3}z^{(i)} =  \mathcal{S} {\times}_1 {U}^{(D)} {\times}_2 {U}^{(C)} {\times}_3 ({U}^{(B)}z^{(i)}).
	\label{bc_eq2tensortucker}
\end{equation}

\subsubsection{TT decomposition}

Given the same sized tensor $\mathcal{W}$, Tensor-Train (TT) decomposition produces two matrices $U^{(D)}$ of size $D\times K_D$ and $U^{(B)}$ of size $K_B \times B$ and a third-order tensor $\mathcal{S}$ of size $K_D \times C \times K_B$, so that

\begin{eqnarray}
\mathcal{W}_{d,c,b} &=& \sum_{k_D=1}^{K_D}\sum_{k_B=1}^{K_B} U^{(D)}_{d,k_D} \mathcal{S}_{k_D,c,k_B} U^{(B)}_{k_B,b},\\
\mathcal{W} &=& \mathcal{S} {\times}_1 {U^{(D)}}^T {\times}_3 U^{(B)}.\label{tt}
\end{eqnarray}

\noindent Substituting Eq.~\ref{tt} into Eq.\ref{bc_eq3tensor}, we obtain the MDL/MTL prediction

\begin{equation}
y = ({U^{(D)}}^T x)\otimes(U^{(B)}z)) \mathcal{S}_{(2)}^T
\label{eq:ytt}
\end{equation}  

\noindent where $\mathcal{S}_{(2)}$ is the mode-2 unfolding of $\mathcal{S}$ which is a $C \times K_D K_B$ matrix, and its transpose $\mathcal{S}_{(2)}^T$ is a matrix of size $K_D K_B \times C$. The weight generating function (Eq.~\ref{bc_eq2tensor}) for Tensor Train decomposition is

\begin{equation}
W^{(i)} = f(z^{(i)}) = \mathcal{W}\times_{3}z^{(i)} = \mathcal{S} {\times}_1 {U^{(D)}}^T {\times}_3 (U^{(B)}z^{(i)}).
\label{bc_eq2tensortt}
\end{equation}

\subsection{Gated Neural Network Architectures}
\label{sec:gnntensor}
We previously showed the connection between matrix factorisation for single-output models, and a two-sided neural network in Section~\ref{sec:twosidenn}. We will next draw the link between tensor factorisation and gated neural network \cite{sigaud2015gatedInventory} architectures. First we recap the  factors used by different tensor (de)composition methods in Table~\ref{tab:factorisationSummary}.

\begin{table}[t]
\centering
\begin{tabular}{c | l l l l}
\hline
Method & \multicolumn{4}{|c}{Factors (Shape)}\\\hline
CP &~ $U^{(D)}$ ($K\times D$) ~  & ~$U^{(C)}$ ($K\times C$)~ & ~$U^{(B)}$ ($K\times B$) & \\
Tucker &~ $U^{(D)}$ ($K_D\times D$) ~ & ~$U^{(C)}$ ($K_C\times C$)~ & ~$U^{(B)}$ ($K_B\times B$) & ~$\mathcal{S}$ ($K_D \times K_C \times K_B$) \\
TT &~ ${U^{(D)}}$ ($D \times K_D$) ~ & & ~$U^{(B)}$ ($K_B\times B$)~ & ~$\mathcal{S}$ ($K_D \times C \times K_B$)\\
\hline
\end{tabular}\caption{Summary of factors used by different tensor (de)composition methods.}\label{tab:factorisationSummary}
\end{table}

To make the connection to neural networks, we need to introduce two new layers:

\begin{description}
	\item[\textbf{Hadamard Product Layer}] Takes as input two equal-length vectors $u$ and $v$ and outputs $[u_1v_1, u_2v_2, \cdots,u_Kv_K]$. It is a deterministic layer that does Hadamard (element-wise) product, where the input size is $K+K$ and output size is $K$.
	\item[\textbf{Kronecker Product Layer}] Takes as input two arbitrary-length vectors $u$ and $v$ and outputs $[u_1v_1, u_2v_1, \cdots, u_{K_u}v_1, u_{1}v_2, \cdots, u_{K_u}v_{K_v}]$. It is a deterministic layer that takes input of size $K_u + K_v$ and returns the size $K_uK_v$ Kronecker product.
	
\end{description}

\begin{figure}
\centering
\subfigure[Single domain learning]{\includegraphics[width=0.44\linewidth]{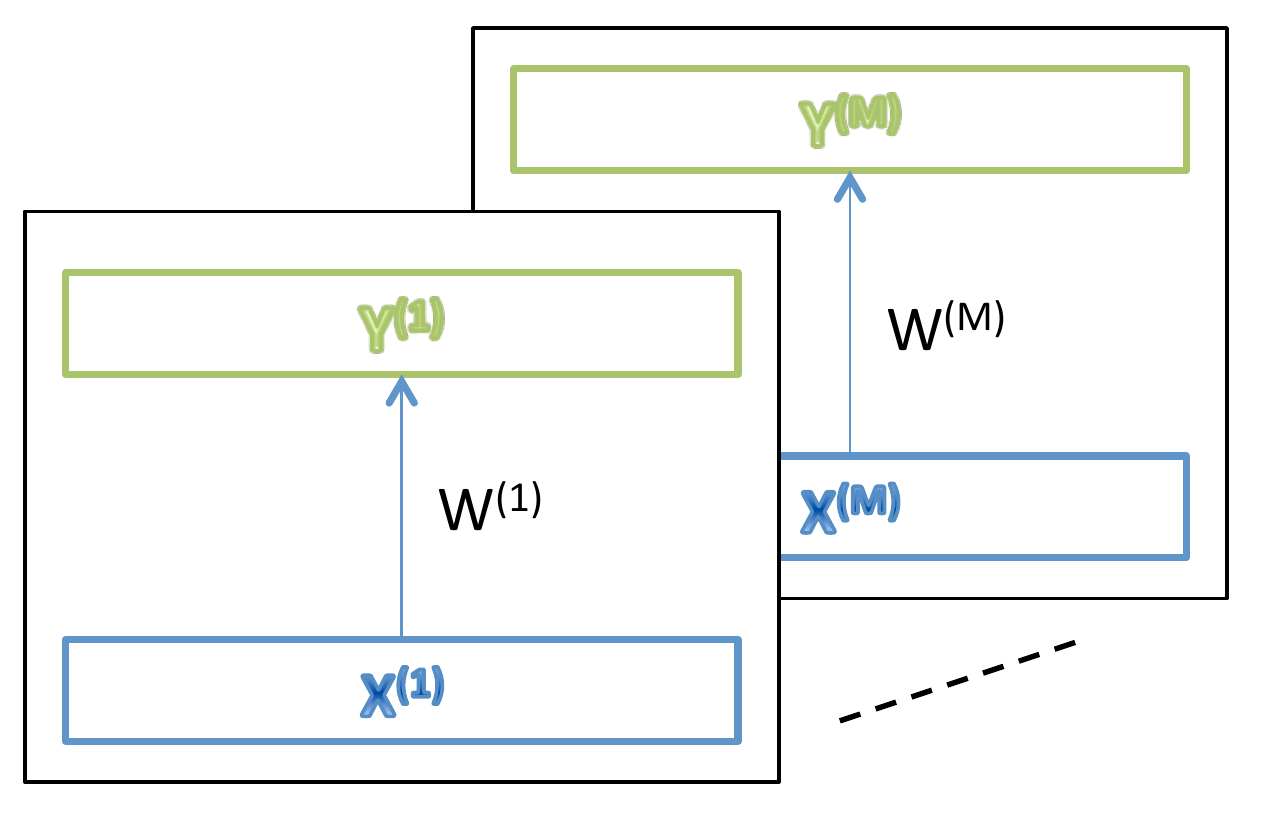}}~~
\subfigure[CP network]{\includegraphics[width=0.44\linewidth]{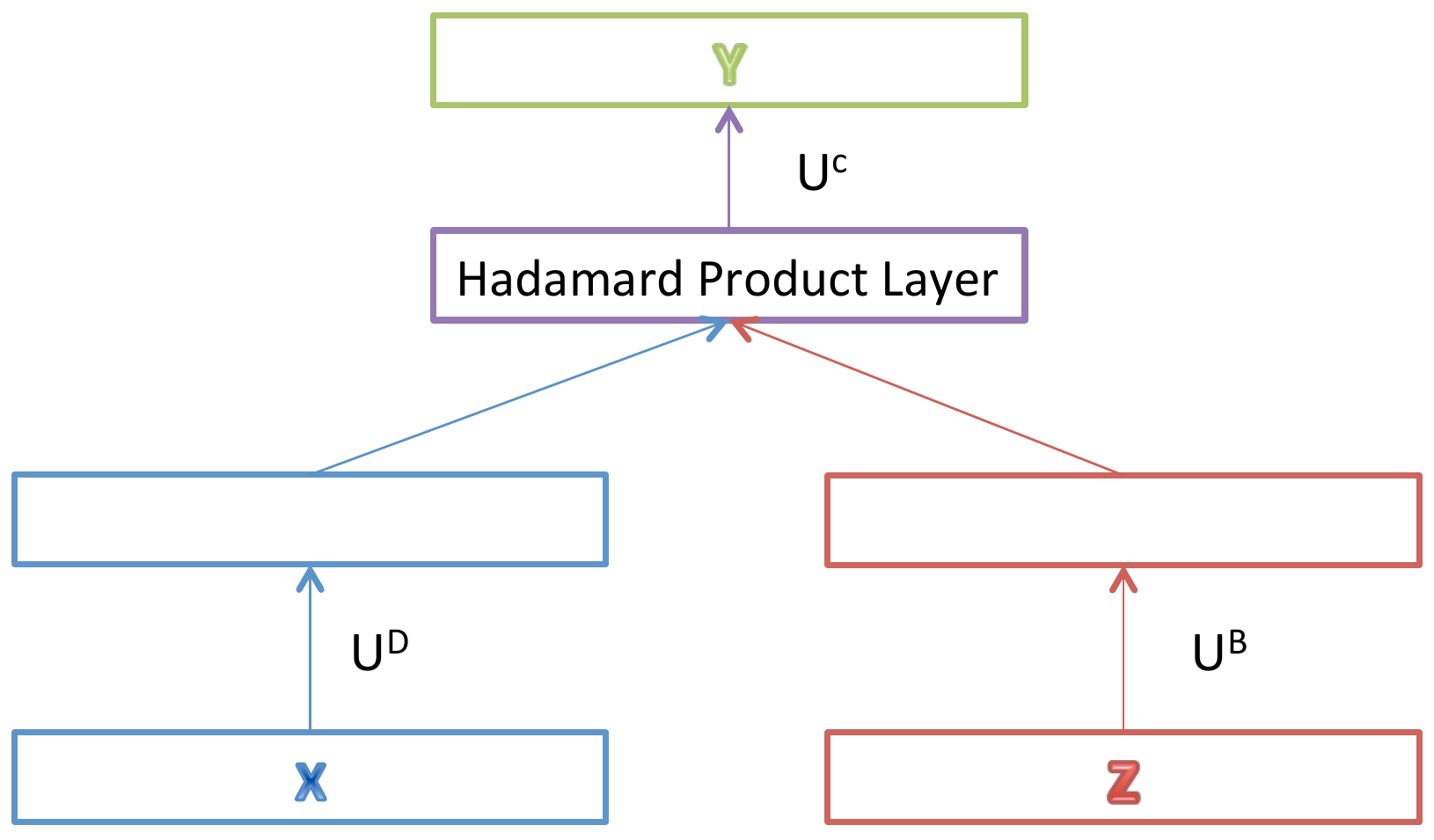}}
\subfigure[TT network]{\includegraphics[width=0.44\linewidth]{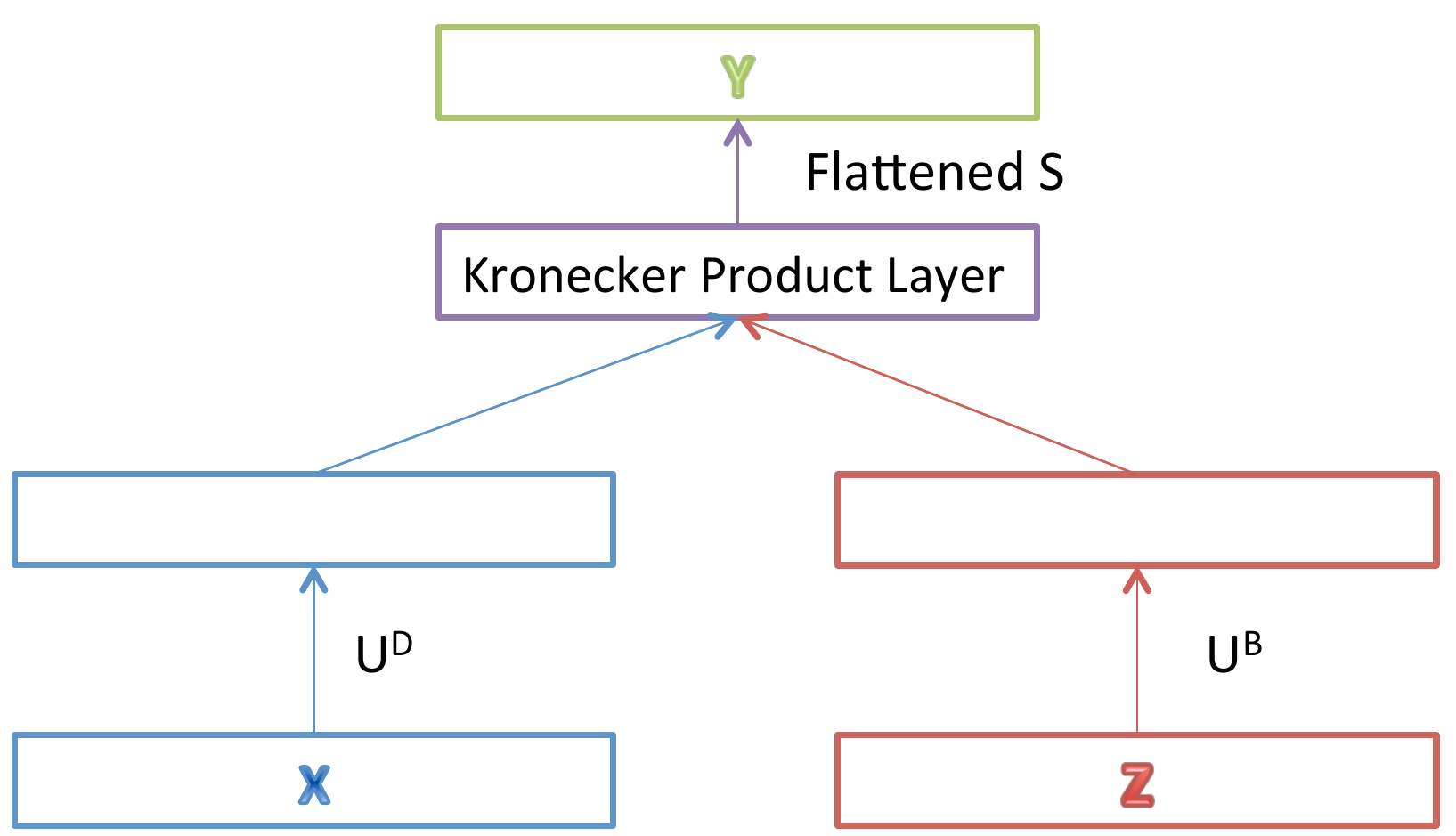}}~~
\subfigure[Tucker network]{\includegraphics[width=0.44\linewidth]{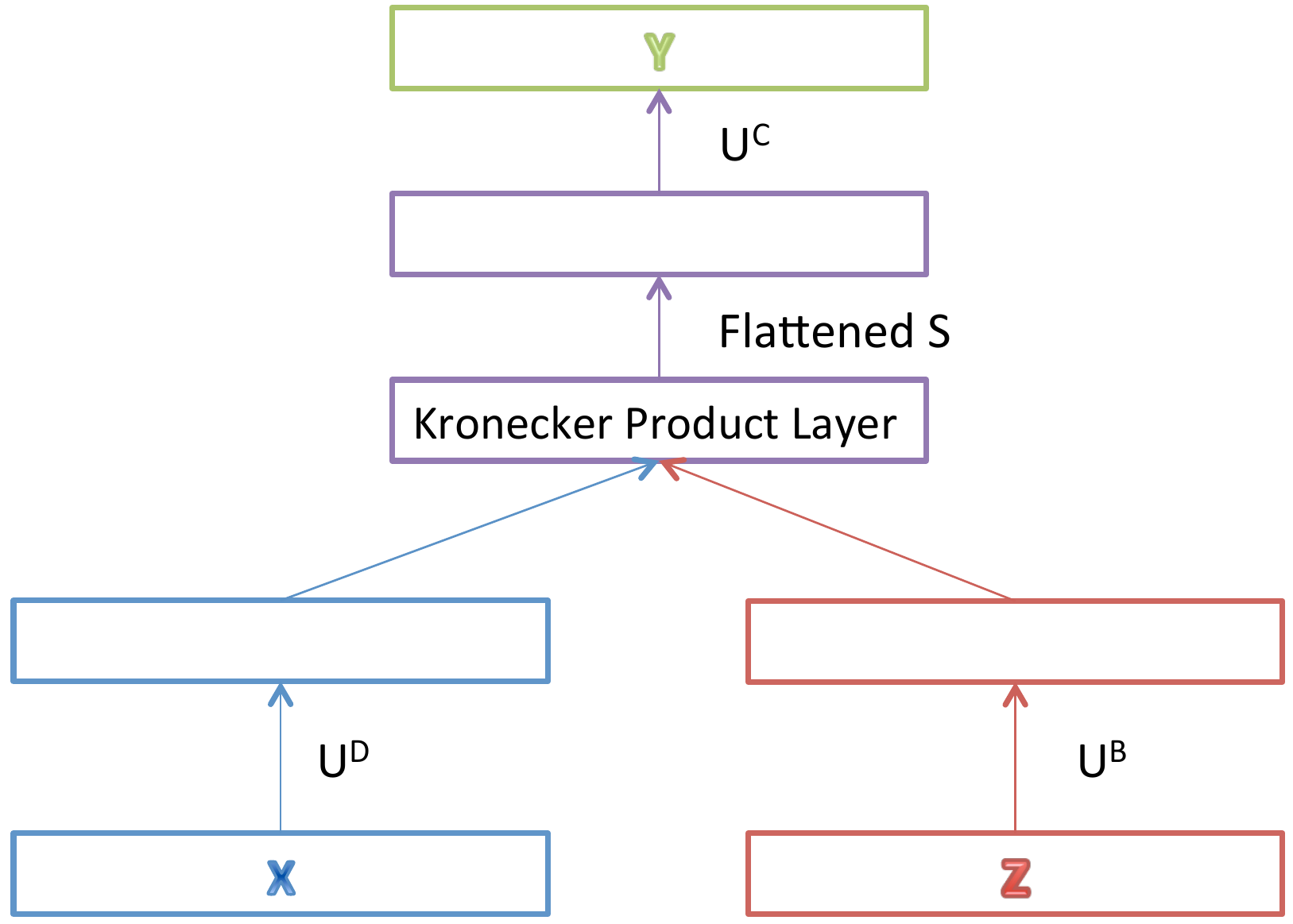}}
\caption{Learning multiple domains independently, versus learning with parametrised neural networks encoding the factorisation assumptions of various tensor decomposition methods.\label{fig:gnntensor}}
\end{figure}

Fig.~\ref{fig:gnntensor} illustrates the approaches to multi-domain learning in terms of NNs. Single domain learning learning of $M$ domains requires $M$ single-layer NNs, each with a $D\times C$ weight matrix (Fig.~\ref{fig:gnntensor}(a)). Considering this set of weight matrices as the corresponding $D\times C\times M$ tensor, we can use the introduced layers to define gated networks (Figs.~\ref{fig:gnntensor}(b)-(d))  that model low-rank versions of this tensor with the corresponding tensor-factorisation assumptions in Eq.~\ref{eq:ycp},~\ref{eq:ytucker}, and~\ref{eq:ytt} and summarised in Tab~\ref{tab:factorisationSummary}. Rather than maintaining a separate NN for each domain as in Fig.~\ref{fig:gnntensor}(a), the networks in Fig.~\ref{fig:gnntensor}(b)-(d) maintain a single NN for all domains. The domain of each instance is signalled to the network via its corresponding descriptor, which the right hand side of the network uses to synthesise the recognition weights accordingly.

We note that we can further unify all three designs, as well as the single-output model proposed in Section~\ref{sec:twosidenn}, by casting them as special cases of the Tucker Network as shown in Table~\ref{tab:allNets}. Thus we can understand all these factorisation-based approaches by their connection to the idea of breaking down the stacked model parameters (matrix $W$ or tensor $\mathcal{W}$) into a smaller number of parameters composing a domain-specific ($U^{(B)}$), task-specific  ($U^{(C)}$) and shared components ($U^{(D)}$). It is important to note however that, despite our model's \emph{factorised representation assumption} in common with tensor decomposition, the way to train our model is not by training a set of models and decomposing them -- in fact matrix/tensor \emph{decomposition} is not used at all. Rather a single Tucker network of Fig.~\ref{fig:gnntensor}(d) is trained end-to-end with backpropagation to minimise the multi-domain/task loss. The network architecture enforces that backpropagation trains the individual factors (Tab~\ref{tab:factorisationSummary}) such that their corresponding tensor \emph{composition} solves the multi-domain-multi-task problem. In summary, our framework can be seen as `discriminatively trained' tensor factorisation, or as a gated neural network, where the NN's weights are dynamically \emph{parametrised} by a second input, the  descriptor $z$.

\begin{table}[t]
\centering
\begin{tabular}{c | c c c c}
\hline
	Method/Factors & $U^{(D)}$ & $U^{(C)}$ & $U^{(B)}$ & $\mathcal{S}$ \\ \hline
	Tucker & $U^{(D)}$ &  $U^{(C)}$ & $U^{(B)}$ & $\mathcal{S}$ \\ 
	CP & $U^{(D)}$ &  $U^{(C)}$ & $U^{(B)}$ & \underline{$K\times K \times K$ Identity Tensor} \\ 
	TT & ${U^{(D)}}^T$ & \underline{$C\times C$ Identity Matrix} & $U^{(B)}$ & $\mathcal{S}$ \\
	Single Output & $P^T$ & \underline{$K\times 1$ All-ones Vector} & $Q$ & \underline{$K\times K \times K$ Identity Tensor} \vspace{0.1em} \\  \hline 
\end{tabular}\caption{Tensor and the matrix-factorisation-based single output (Sec~\ref{sec:twosidenn}, \cite{YangICLR15Unified}) networks as special cases of the Tucker Network.  \underline{Underlined variables} are constant rather than learned parameters.}\label{tab:allNets}
\end{table}

\section{Experiments}

We explore three sets of experiments on object recognition (Section~\ref{sec:officeMDL}), surveillance image analysis (Section~\ref{sec:survilliance}) and person recognition/soft-biometrics (Section~\ref{sec:gait}). The first recognition experiment follows the conventional setting of domains/tasks as atomic entities, and the latter experiments explore the potential benefits of informative domain descriptors, including zero-shot domain adaptation.

\vspace{0.1cm}\noindent\textbf{Implementation}\quad We implement our framework with TensorFlow \cite{tensorflow2015}, taking the neural network interpretation of each method,  thus allowing easy optimisation with SGD-based backpropagation. We use hinge loss for the binary classification problems and (categorical) cross-entropy loss for the multi-class classification problems.

\subsection{Multi-domain Multi-task Object Recognition}\label{sec:officeMDL}

In this section we assume conventional atomic domains (so domain descriptors are simply indicators rather than distributed codes), but explore a multi-domain multi-task (MDMTL) setting. Thus there is a multi-class problem within each domain, and our method (Section~\ref{sec:multileouput}) exploits information sharing across both domains and tasks. To deal with multi-class recognition within each domain, it generalises existing vector-valued MTL/MDL methods, and implements a matrix-valued weight generating function parametrised by a low-rank tensor (Fig.~\ref{fig:gnntensor}).

\begin{figure}[t]
	\centering
	\includegraphics[width=0.95\columnwidth]{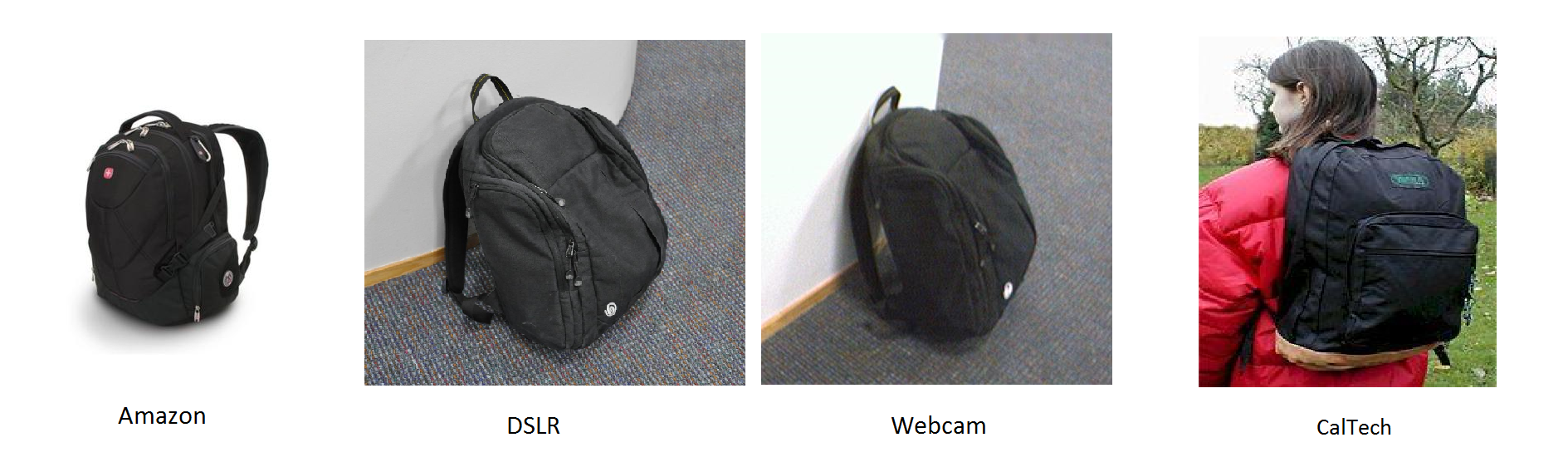}
	\caption{Illustration of the domains in the office dataset. An image of a backpack collected from four different sources.}
	\label{figoffice}
\end{figure}

\vspace{0.1cm}\noindent\textbf{Datasets}\quad We first evaluate the multi-domain multi-task setting using the well-known office dataset \cite{SaenkoECCV10Adapting}. Office includes three domains (data sources): \emph{Amazon}: images downloaded from Amazon, \emph{DSLR} high-quality images captured by digital camera, \emph{webcam} low-quality images captured by webcam. For every domain, there are multiple classes of objects to recognise, e.g., keyboard, mug, headphones. 
In addition to the original Office dataset, add a 4th domain: Caltech-256 \cite{caltech256}, as suggested by \cite{GongCVPR12Geodesic}. Thus we evaluate recognising 10 classes in common the four domains. See Fig.~\ref{figoffice} for an illustration.  The feature is the $800$-dimension SURF feature \cite{Bay2008surf}. As suggested by \cite{GongCVPR12Geodesic}, we pre-process the data by normalising the sum of each instance's feature vector to one then applying a z-score function.

\vspace{0.1cm}\noindent\textbf{Settings}\quad We compare the three proposed method variants: CP, Tucker, and TT-Networks with two baselines. \textbf{SDL:} training each domain independently and \textbf{Aggregation}: ignoring domains and training an aggregate model for all data. For these two baselines, we use a vanilla feed-forward neural network without hidden layers thus there are no hyper-parameters to tune. For our methods, the tensor rank(s), i.e., $K$ for CP-Network, ($K_D$, $K_C$, $K_B$) for Tucker-Network, and ($K_D$, $K_B$) for TT-network are chosen by 10-fold cross validation. The grids of $K_D$, $K_C$, and $K_B$ are respectively $[16,64,256]$, $[2,4,8]$, and $[2,4]$. The multi-class recognition error rate at 9 increasing training-testing-ratios ($10\%, 20\% \dots 90\%$) is computed, and for each training-testing-ratio, we repeat the experiment 10 times with random splits.

\begin{figure}[t]
	\centering
	\includegraphics[width=0.8\columnwidth]{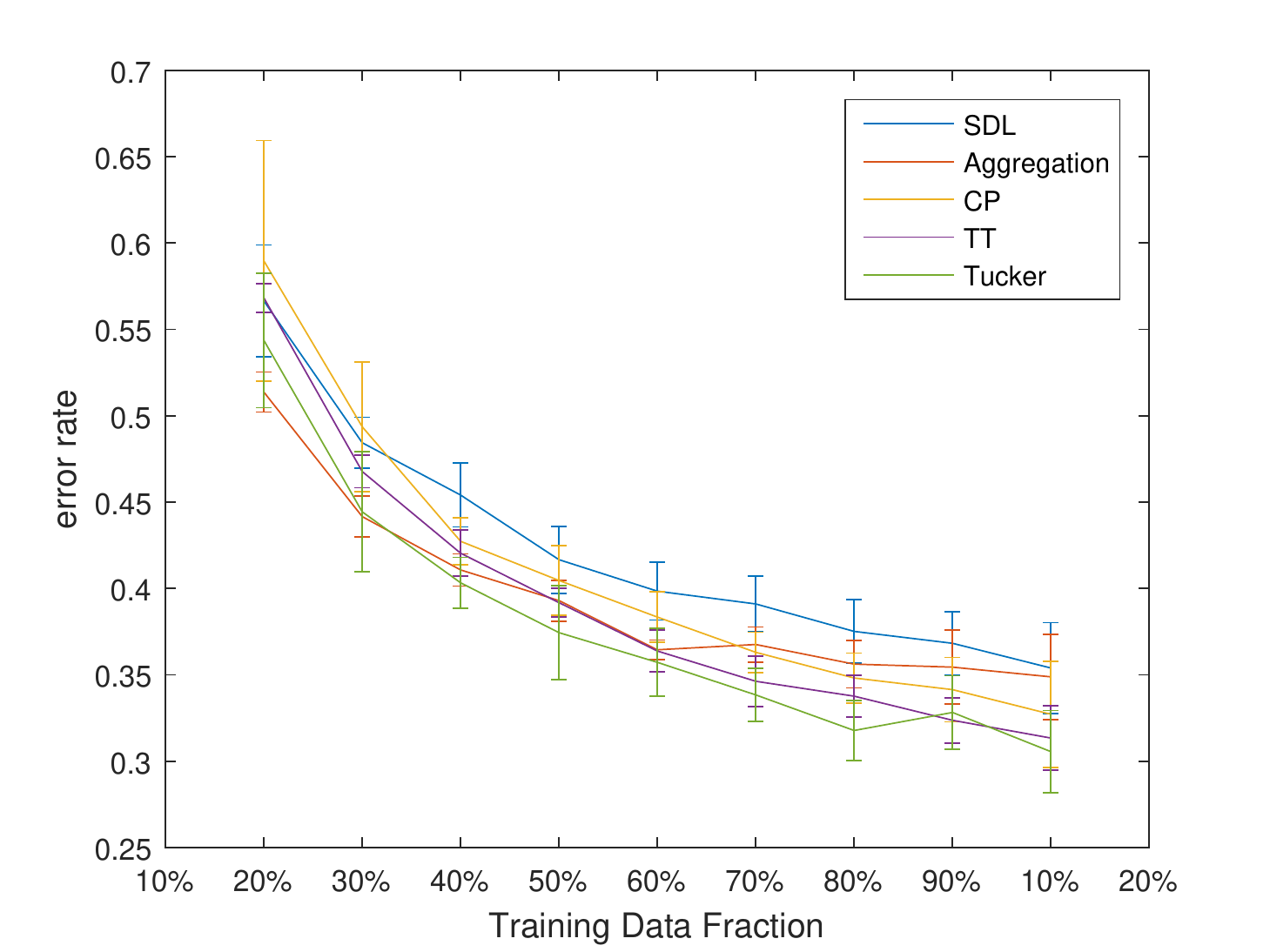}
	\caption{Office Dataset: Error mean and std. dev. recognising $C=10$ classes across $M=4$ domains. Comparison of three parametrised neural networks with SDL and Aggregation baselines.}
	\label{figofficere}
\end{figure}

\vspace{0.1cm}\noindent\textbf{Results and Analysis}\quad The result is shown in Fig.~\ref{figofficere}. We can see that the proposed methods perform well compared to \emph{SDL}. When the training data is extremely small, \emph{Aggregation} is a reasonably good choice as the benefit of more data outweighs the drawback of mixing domains. However, the existence of domain bias eventually prevents a single model from working well for all domains. Our proposed methods can produce different models for the various domains so they generally outperform the baselines. Tucker-and TT-Network are better than CP-Network because of their greater flexibility on choosing more than one tensor rank. However as a drawback, this also introduces more hyper-parameters to tune.

\subsection{Surveillance Image Classification}\label{sec:survilliance}
\begin{figure}[t]
	\centering
	\fbox{
		\includegraphics[width = 0.30\linewidth]{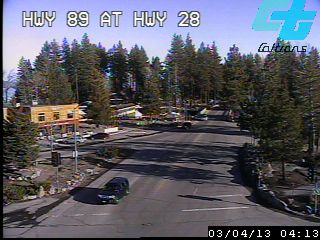}
		\includegraphics[width = 0.30\linewidth]{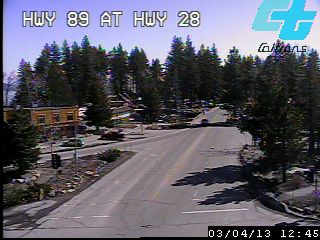}
		\includegraphics[width = 0.30\linewidth]{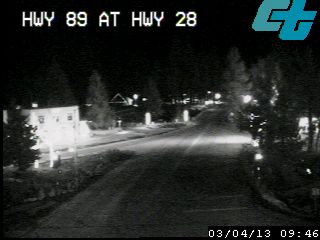}
	}
	\\\vspace{0.1cm}
	\fbox{
		\includegraphics[width = 0.30\linewidth]{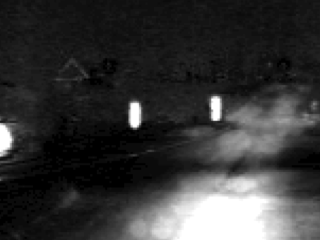}
		\includegraphics[width = 0.30\linewidth]{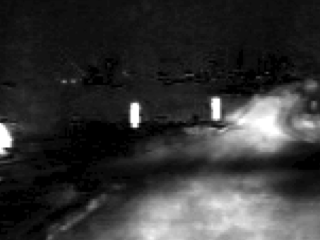}
		\includegraphics[width = 0.30\linewidth]{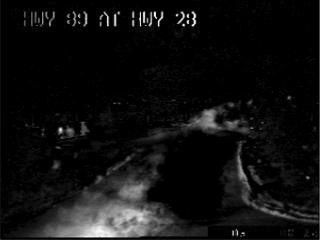}
	}\\\vspace{0.1cm}
	\caption{Illustration of domain factors in surveillance car recognition task.  Above: Example frames illustrating day/night domain factor. (Left: With car. Middle and right: No cars.) Below: Activity map illustration of weekday/weekend domain factor. (Left: Weekday, Middle: Weekend: Right: Weekday-Weekend difference.)}
	\vspace{-0.1cm}
	\label{fig:surveillance}
\end{figure}

In surveillance video analysis, there are many covariates such as season, workday/holiday and day/night. Each of these affects the distribution of image features, and thus introduces domain shift. Collecting the potentially years of training data required to train a  single general model is both expensive and suboptimal (due to ignoring domain shift, and treating all data as a single domain). 
Thus in this section we explore the potential for multi-domain learning with distributed domain descriptors (Section~\ref{sec:singleouput}) to improve performance by modelling the factorial structure in the domain shift. 
Furthermore, we demonstrate the potential of ZSDA  to adapt a system to apply in a new set of conditions for which no training data exists. 

\vspace{0.1cm}\noindent\textbf{Data}\quad We consider the surveillance image classification task proposed by \cite{hoffman2014continuousDA}. This is a binary classification of each frame in a 12-day surveillance stream as being empty or containing cars. Originally, \cite{hoffman2014continuousDA} investigated continuous domains (which can be seen as a 1-dimensional domain descriptor containing time-stamp). To explore a richer domain descriptor, we use a slightly different definition of domains, considering instead  weekday/weekend and day/night as domain factors, \textcolor{black}{generating $2\times2=4$ distinct domains, each encoded by a 2-of-4 binary domain descriptor}.  Figure~\ref{fig:surveillance}(top) illustrates the more obvious domain factor: day/night. This domain-shift induces a larger image change than the task-relevant presence or absence of a car.

\vspace{0.1cm}\noindent\textbf{Settings}\quad We use the 512 dimensional GIST feature for each frame provided by \cite{hoffman2014continuousDA}. We perform two experiments: \emph{Multi-domain learning}, and \emph{zero-shot domain adaptation}. For MDL, we split all domains' data into half training and half testing, and repeat for 10 random splits. We use our single-output network (Fig.~\ref{fig:model}, Tab~\ref{tab:allNets} bottom row) with a distributed domain descriptor for two categories with two states (i.e., same descriptor as Fig.~\ref{fig:ddDistrib}, right). The baselines are: (i) \textbf{SDL}: train an independent model for each domain (ii) \textbf{Aggregation}: to train a single model covering all domains (ii) \textbf{Multi-Domain I}:   a multi-domain model with low-rank factorisation of $\tilde{W}$ and one-hot encoding of domain descriptor (in this case, $Z$ is an identity matrix thus $\tilde{W}=WZ=W$ -- this roughly corresponds to our reimplementation of \cite{daume2012gomtl}), and (iv) \textbf{Multi-Domain II}:  a factorised multi-domain model with one-hot + constant term encoding (this is in fact the combination the sharing structure and factorisation proposed in \cite{Evgeniou2004} and \cite{daume2012gomtl} respectively). 

For ZSDA, we do leave-one-domain-out cross-validation: holding out one of the four domains for testing, and using the observed three domains' data for training. Although the train/test splits are not random, we still repeat the procedure 10 times to reduce randomness of the SGD optimisation.  Our model is constructed on the fly for the held-out domain based on its semantic descriptor. As a baseline, we train an aggregated model from all observed domains' data, and apply it  directly to the held-out domain (denoted as \textbf{Direct}). We set our rank hyper parameter via the heuristic $K=\frac{D}{\log(D)}$. We evaluate the the mean and standard deviation of error rate.

\vspace{0.1cm}\noindent\textbf{Results and Analysis}\quad The results shown in Table~\ref{tab:surveillance} demonstrate that our proposed method outperforms alternatives in both MDL and ZSDA settings. For MDL we see that training a per-domain model and ignoring domains altogether perform similarly (SDL vs Aggregation). By introducing more sharing structure, e.g., Multi-Domain I is built with low-rank assumption, and Multi-Domain II further assumes that there is a globally shared factor, the multi-domain models  clearly improve performance. Finally our full method performs notably better than the others because it can benefits from both low-rank modelling and also exploiting the structured information in the distributed encoding of domain semantic descriptor.

In ZSDA, our proposed method also clearly outperforms the baseline of directly training on all the source domains. What information is our model able to exploit to achieve this? One cue is that various directions including right turn are common on weekends and weekdays are primarily going straight (illustrated in Figure~\ref{fig:surveillance}(below) by way of an activity map). This can, e.g., be learned from the weekend-day domain, and transferred to the held-out weekend-night domain because the domain factors inform us that those two domains have the weekend factor in common.

\begin{table}[t]
\label{tab:surveillance}
\begin{center}
\caption{Surveillance Image Classification. Mean error rate (\%) and standard deviation.}
\scalebox{0.85}{
\begin{tabular}{c | c c c c c }
\hline
MDL Experiment & SDL & Aggregation & Multi-Domain I & Multi-Domain II & Parametrised NN \\ \hline
Err. Rate & 10.82 $\pm$ 3.90 & 11.05 $\pm$ 2.73 & 10.00 $\pm$ 0.90  & 8.86 $\pm$ 0.79 & \textbf{8.61 $\pm$0.51 }\\ \hline
\end{tabular}}
\newline
\vspace*{1em}
\newline
\scalebox{0.85}{
\begin{tabular}{c|  c c}
\hline
ZSDA Experiment & Direct &  Parametrised NN \\ \hline
Err. Rate &12.04 $\pm$0.11 &  \textbf{9.72 $\pm$0.08}\\ \hline
\end{tabular}}
\end{center}
\vspace{-0.6cm}
\end{table}

\subsection{Gait-based Soft-Biometrics and Recognition}\label{sec:gait}

\begin{figure}[t]
\centering
\includegraphics[width=0.80\columnwidth]{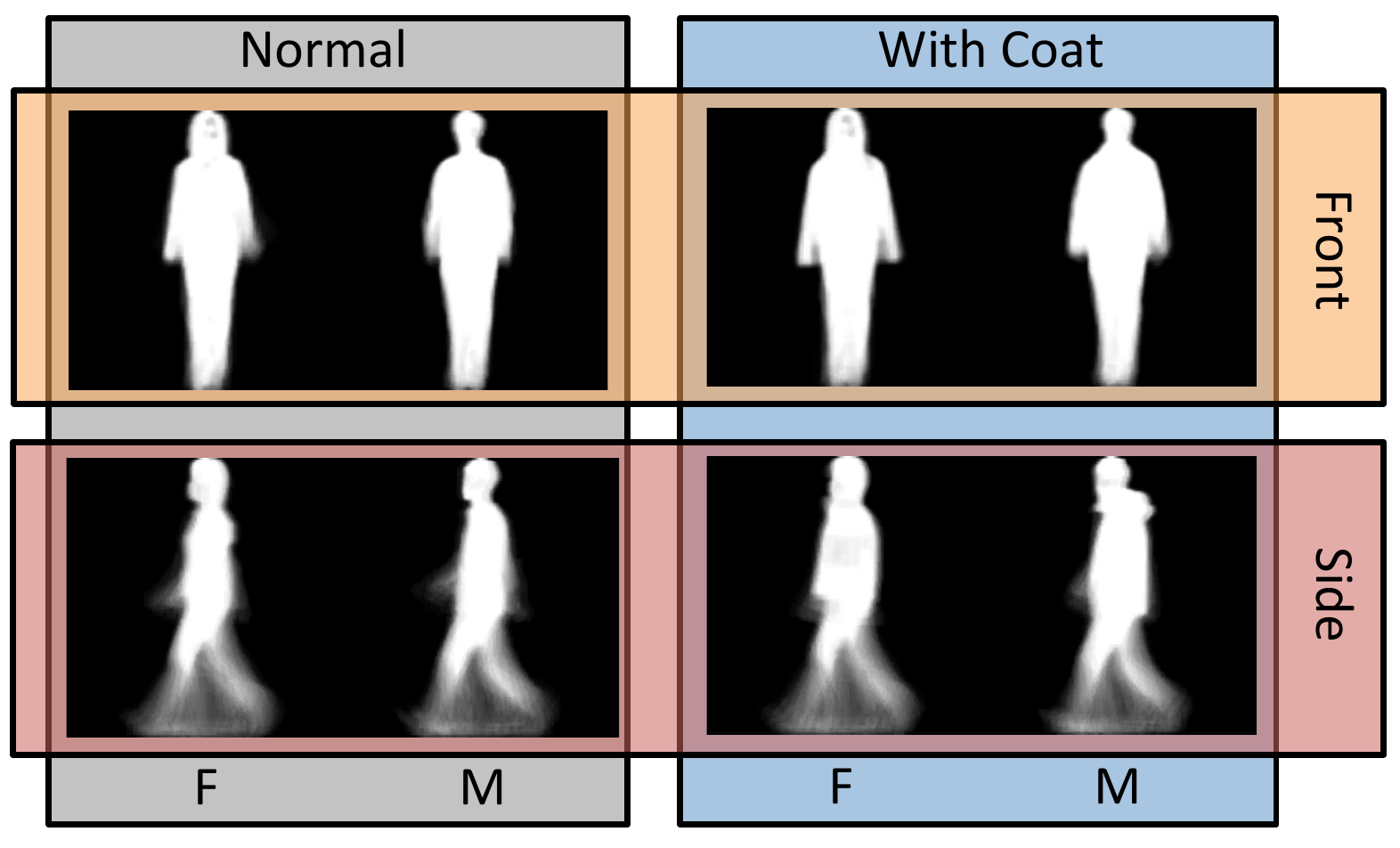}
\caption{Example gait images illustrating independent domain factors.}
\label{figdomainIllustration}
\end{figure}

Gait-based person and soft biometric recognition are desirable capabilities due to not requiring subject cooperation \cite{ZhengICIP2011Robust}. However they are challenging especially where there are multiple covariates such as viewing angle and accessory status (e.g., object carrying). Again training a model for every covariate combination is infeasible, and conventional domain adaptation is not scalable as the number of resulting domains grows exponentially with independent domain factors. In contrast, zero-shot domain adaptation could facilitate deploying a camera with a calibration step to specify covariates such as view-angle, but no data collection or re-training.

\noindent\textbf{Data}\quad We investigate applying our framework to this setting using the CASIA gait analysis dataset \cite{ZhengICIP2011Robust} with data from 124 individuals under 11 viewing angles. Each person has three situations: normal (`nm'), wearing overcoat (`cl') and carrying a bag (`bg'). This naturally forms $3\times 11=33$ domains. We extract Gait Energy Image (GEI) features, followed by PCA reduction to 300 dimensions.

\noindent\textbf{Settings}\quad We consider two gait analysis problems: (i) Soft-biometrics: Female/Male classification and (ii) Person verification/matching. For matching each image pair $x_i$ and $x_j$, generates a pairwise feature vector by $x_{ij}=|x_i-x_j|$. The objective is to learn a binary verification classifier on $x_{ij}$ to predict if two images are the same person or not. All experiment settings (baseline methods, training/testing splits, experiments repeats, and the choice of hyper-parameter) are the same as in Section~\ref{sec:survilliance}, except that for the verification problem we build a balanced (training and testing) set of positive/negative pairs by down-sampling negative pairs.

\vspace{0.1cm}\noindent\textbf{Results and Analysis}\quad Figure~\ref{figdomainIllustration} illustrates the nature of the domain factors here, where the cross-domain variability is again large compared to the cross-class variability. Our framework uniquely models the fact that each domain factor (e.g., view angle and accessory status) can occur independently. The results shown in Tables~\ref{gait:biometric} and \ref{gait:reid} demonstrate the same conclusions -- that explicitly modelling MDL structure improves performance (Multi-domain I and II improve on SDL and Aggregation), with our most general method performing best overall.

\begin{table}[h]
\caption{Gait: Male/Female Biometrics. Error Rate (\%) and Standard Deviation} \label{gait:biometric}
\begin{center}
\scalebox{0.85}{
\begin{tabular}{c | c c c c c }
\hline
MDL Experiment & SDL & Aggregation & Multi-Domain I & Multi-Domain II & Parametrised NN \\ \hline
Err. Rate & 2.35 ($\pm$0.20) & 2.62 ($\pm$0.18) & 2.25 ($\pm$0.20) & 2.07 ($\pm$ 0.15) & \textbf{1.64 ($\pm$0.14)}\\ \hline
\end{tabular}}
\newline
\vspace*{1em}
\newline
\scalebox{0.85}{
\begin{tabular}{c|  c c}
\hline
ZSDA Experiment & Direct &  Parametrised NN \\ \hline
Err. Rate &3.01 ($\pm$0.08) & \textbf{2.19 ($\pm$0.05)} \\\hline
\end{tabular}
}
\end{center}
\end{table}

\begin{table}[h]
\caption{Gait: Person Verification. Error Rate(\%) and Standard Deviation}\label{gait:reid}
\begin{center}
\scalebox{0.85}{
\begin{tabular}{c | c c c c c }
\hline
MDL Experiment & SDL & Aggregation & Multi-Domain I & Multi-Domain II & Parametrised NN \\ \hline
Err. Rate & 23.30 ($\pm$0.28) & 24.62 ($\pm$0.32) & 22.18 ($\pm$0.25) & 21.15 ($\pm$ 0.13) & \textbf{19.36 ($\pm$0.09)}\\ \hline
\end{tabular}
}
\newline
\vspace*{2em}
\newline
\scalebox{0.85}{
\begin{tabular}{c|  c c}
\hline
ZSDA Experiment& Direct &  Parametrised NN \\ \hline
Err. Rate &26.93 ($\pm$0.10) & \textbf{23.67 ($\pm$0.11)}\\ \hline
\end{tabular}
}
\vspace{-0.6cm}
\end{center}
\end{table}

\section{Conclusion}

In this chapter, we discussed multi-domain learning, a bi-directional generalisation of domain adaptation and zero-shot domain adaptation, an alternative to domain-generalisation approaches. We introduced a semantic domain/task descriptor to unify various existing multi-task/multi-domain algorithms within a single matrix factorisation framework. To go beyond the single output problems considered by prior methods, we generalised this framework to tensor factorisation, which allows knowledge sharing for methods parametrised by matrices rather than vectors. This allows multi-domain learning for multi-output problems or simultaneous multi-task-multi-domain learning. All these approaches turn out to have equivalent interpretations as neural networks, which allow easy implementation and optimisation with existing toolboxes. Promising lines of future enquiry include extending this framework for end-to-end learning in convolutional neural networks \cite{YangX16tensor}, tensor rank-based regularisation \cite{YangX16tensortracenorm}, and applying these ideas to solve practical computer vision problems.


\end{document}